\documentclass[11pt,a4paper]{article}
\usepackage[hyperref]{emnlp2020}
\usepackage{times}
\usepackage{latexsym}

\usepackage{microtype}
\usepackage{amsmath}
\usepackage{amssymb}

\usepackage{graphicx}
\usepackage{comment}
\usepackage{array}
\usepackage{mathtools}
\usepackage[linguistics]{forest}
\usepackage{xcolor}
\usepackage{comment}
\usepackage{soul}
\usepackage{tikz-dependency}

\aclfinalcopy 

\title{Event Detection: Gate Diversity and Syntactic Importance Scores \\for Graph Convolution Neural Networks}

\author{ Viet Dac Lai\textsuperscript{\rm 1}, Tuan Ngo Nguyen\textsuperscript{\rm 1} and Thien Huu Nguyen\textsuperscript{\rm 1,2} \\
\textsuperscript{\rm 1} Department of Computer and Information Science, University of Oregon,
\\Eugene, OR 97403, USA\\
\textsuperscript{\rm 2} VinAI Research, Vietnam\\
  \texttt{\{vietl,tnguyen,thien\}@cs.uoregon.edu} \\
}

\date{}

\begin{document}
\maketitle
\begin{abstract}
Recent studies on event detection (ED) have shown that the syntactic dependency graph can be employed in graph convolution neural networks (GCN) to achieve state-of-the-art performance. However, the computation of the hidden vectors in such graph-based models is agnostic to the trigger candidate words, potentially leaving irrelevant information for the trigger candidate for event prediction. In addition, the current models for ED fail to exploit the overall contextual importance scores of the words, which can be obtained via the dependency tree, to boost the performance. In this study, we propose a novel gating mechanism to filter noisy information in the hidden vectors of the GCN models for ED based on the information from the trigger candidate. We also introduce novel mechanisms to achieve the contextual diversity for the gates and the importance score consistency for the graphs and models in ED. The experiments show that the proposed model achieves state-of-the-art performance on two ED datasets. 

\end{abstract}

\section{Introduction}

Event Detection (ED) is an important task in Information Extraction of Natural Language Processing. The main goal of this task is to identify event instances presented in text. Each event mention is associated with a word or a phrase, called an event trigger, which clearly expresses the event \cite{walker2006ace}. The event detection task, precisely speaking, seeks to identify the event triggers and classify them into some types of interest. For instance, consider the following sentences:

(1) \textit{They'll be \textbf{fired} on at the crossing}.

(2) \textit{She is on her way to get \textbf{fired}}.

An ideal ED system should be able to recognize the two words ``{\it fired}'' in the sentences as the triggers of the event types ``Attack'' (for the first sentence) and ``End-Position'' (for the second sentence).


The dominant approaches for ED involve deep neural networks to learn effective features for the input sentences, including separate models \cite{Chen:15} and joint inference models with event argument prediction \cite{Nguyen:19}. Among those deep neural networks, graph convolutional neural networks (GCN) \cite{Kipf:17} have achieved state-of-the-art performance due to the ability to exploit the syntactic dependency graph to learn effective representations for the words \cite{nguyen:18:graph,liu2018jointly,yan2019event}. However, two critical issues should be addressed to further improve the performance of such models.

First, given a sentence and a trigger candidate word, the hidden vectors induced by the current GCN models are not yet customized for the trigger candidate. As such, the trigger-agnostic representations in the GCN models might retain redundant/noisy information that is not relevant to the trigger candidate. As the trigger candidate is the focused word in the sentence, that noisy information might impair the performance of the ED models. To this end, we propose to filter the noisy information from the hidden vectors of GCNs so that only the relevant information for the trigger candidate is preserved. In particular, for each GCN layer, we introduce a gate, computed from the hidden vector of the trigger candidate, serving as the irrelevant information filter for the hidden vectors. Besides, as the hidden vectors in different layers of GCNs tend to capture the contextual information at different abstract levels, we argue that the gates for the different layers should also be regulated to exhibit such abstract representation distinction. Hence, we additionally introduce a novel regularization term for the overall loss function to achieve these distinctions for the gates.

Second, the current GCN models fail to consider the overall contextual importance scores of every word in the sentence. In previous GCN models, to produce the vector representation for the trigger candidate word, the GCN models mostly focus on the closest neighbors in the dependency graphs \cite{nguyen:18:graph,liu2018jointly}. However, although the non-neighboring words might not directly carry useful context information for the trigger candidate word, we argue that their overall importance scores/rankings in the sentence for event prediction can still be exploited to provide useful training signals for the hidden vectors in ED. In particular, we propose to leverage the dependency tree to induce a \textbf{graph-based} importance score for every word based on its distance to the trigger candidate. Afterward, we propose to incorporate such importance scores into the ED models by encouraging them to be consistent with another set of \textbf{model-based} importance scores that are computed from the hidden vectors of the models. Based on this consistency, we expect that graph-based scores can enhance the representation learning for ED. In our experiments, we show that our method outperforms the state-of-the-art models on the benchmark datasets for ED.

\section{Related Work}

Prior studies on ED involve handcrafted feature engineering for statistical models \cite{Ahn:06,Ji:08,Hong:11,Li:13,Mitamura:15} and deep neural networks, e.g., CNN \cite{Chen:15,Chen:17,Nguyen:15:event,Nguyen:16g}, RNN \cite{Nguyen:16a:joint,Jagannatha:16,Feng:16}, attention mechanism \cite{Liu:17,Chen:18},
contextualized embeddings \cite{yang2019exploring}, and adversarial training \cite{wang2019adversarial}. The last few years witness the success of graph convolutional neural networks for ED \cite{nguyen:18:graph,liu2018jointly,veyseh:19b,yan2019event} where the dependency trees are employed to boost the performance. However, these graph-based models have not considered representation regulation for GCNs and exploiting graph-based distances as we do in this work. 

\section{Model}

\textbf{Task Description}: The goal of ED consists of identifying trigger words (\textbf{trigger identification}) and classifying them for the event types of interest (\textbf{event classification}). Following the previous studies \cite{Nguyen:15:event}, we combine these two tasks as a single multi-way classification task by introducing a \textit{None} class, indicating non-event. Formally, given a sentence $X=[x_1,x_2,\ldots,x_n]$ of $n$ words, and an index $t$ ($1\leq t \leq n$) of the trigger candidate $x_t$, the goal is to predict the event type $y^*$ for the candidate $x_t$. Our ED model consists of three modules: (1) Sentence Encoder, (2) GCN and Gate Diversity, and (3) Graph and Model Consistency.

\textbf{Sentence Encoder}: We employ the pre-trained BERT \cite{devlin2019bert} to encode the given sentence $X$. 
In particular, we create an input sequence of $[[CLS], x_1, \cdots, x_n, [SEP], x_t, [SEP]]$ where $[CLS]$ and $[SEP]$ are the two special tokens in BERT. The word pieces, tokenized from the words, are fed to BERT to obtain the hidden vectors of all layers. We concatenate the vectors of the top $M$ layers to obtain the corresponding hidden vectors for each word piece, where $M$ is a hyper-parameter. Then, we obtain the representation of the sentence $E=\{e_1, \cdots, e_n\}$ in which the vectors $e_i$ of $x_i$ is the average of layer-concatenated vectors of its word pieces. Finally, we feed the embedding vectors in $E$ to a bidirectional LSTM, resulting in a sequence of hidden vectors $h^0=\{h^0_1, \cdots, h^0_n\}$.

\newcommand{\gp}{\mathcal{G}}
\newcommand{\vp}{\mathcal{V}}
\newcommand{\ep}{\mathcal{E}}
\textbf{GCN and Gate Diversity}: To apply the GCN model, we first build the sentence graph $\gp=(\vp, \ep)$ for $X$ based on its dependency tree, where $\vp, \ep$ are the sets of nodes and edges, respectively. $\vp$ has $n$ nodes, corresponding to the $n$ words $X$. Each edge $(x_i, x_j)$ in $\ep$ amounts to a directed edge from the head $x_i$ to the dependent $x_j$ in the dependency tree. Following \cite{Marcheggiani:17}, we also include the opposite edges of the dependency edges and the self-loops in $\ep$ to improve the information flow in the graph. 




Our GCN module contains $L$ stacked GCN layers \cite{Kipf:17}, operating over the sequence of hidden vectors $h^0$. The hidden vector $h_i^l$ ($1\leq i \leq n, 1\leq l\leq L)$ of the word $x_i$ at the $l$-th layer is computed by averaging the hidden vectors of neighboring nodes of $x_i$ at the $(l-1)$-th layer. Formally, $h^l_i$ is computed as follow:
\begin{equation}
    h_i^l=\text{ReLU}\left(W^l\sum_{(x_i,x_j)\in \ep}^{}\frac{h^{l-1}_j}{|\{x_j\}|}\right)
\end{equation}
where $W^l$ is a learnable weight of the GCN layer.

The major issue of the current GCN for ED is that its hidden vectors $h^l_i$ are induced without special awareness of the trigger candidate $x_t$. This might result in irrelevant information (for the trigger word candidate) in the hidden vectors of GCNs for ED, thus hindering further performance improvement. To address this problem, we propose to filter that unrelated information by introducing a gate for each GCN layer. The vector $g^l$ for the gate at the $l$-th layer is computed from the embedding vector $e_t$ of the trigger candidate: $g^l=\sigma(W_g^le_t)$  where $W_g^l$ are learnable parameters for the $l$-th layer. Then, we apply these gates over the hidden vectors of the corresponding layer via the element-wise product, resulting in the filtered vectors: $m^l_i=g^l\circ h^l_i$.

As each layer in the GCN module has access to a particular degree of neighbors, the contextual information captured in these layers is expectedly distinctive. Besides, the gates for these layers control which information is passed through, therefore, they should also demonstrate a certain degree of contextual diversity. To this end, we propose to encourage the distinction among the outcomes of these gates once they are applied to the hidden vectors in the same layers. Particularly, starting with the hidden vectors $h^l$ of of the $l$-layer, we apply the gates $g^k$ (for all $(1\leq k\leq L)$) to the vectors in $h^l$, which results in a sequence of filtered vectors $\bar{m}^{k,l}_i=g^k \circ h^l_i$. Afterward, we aggregate the filtered vectors obtained by the same gates using max-pooling: $\bar{m}^{k,l}=\text{max\_pool}(\bar{m}^{k,l}_1, \cdots, \bar{m}^{k,l}_n)$. To encourage the gate diversity, we enforce vector separation between $\bar{m}^{l,l}$ with all the other aggregated vectors from the same layer $l$ (i.e., $\bar{m}^{k,l}$ for $k\ne l$). As such, we introduce the following cosine-based regularization term $\mathcal{L}_{GD}$ (for Gate Diversity) into the overall loss function:
\begin{equation}
\mathcal{L}_{GD} = \frac{1}{L(L-1)}\sum_{l=1}^L\sum_{k=l+1}^{L} \text{cosine}(\bar{m}^{l,l}, \bar{m}^{l,k})
\end{equation}
Note that the rationale for applying the gates $g^k$ to the hidden vectors $h^l$ for the gate diversity is to ground the control information in the gates to the contextual information of the sentence in the hidden vectors to facilitate meaningful context-based comparison for representation learning in ED.

\textbf{Graph and Model Consistency}: As stated above, we seek to supervise the model using the knowledge from the dependency graph. Inspired by the contextual importance of the neighboring words for the event prediction of the trigger candidate $x_t$, we compute the \textbf{graph-based importance scores} $P={p_1,\cdots, p_n}$ in which $p_i$ is the negative distance from the word $x_i$ to the trigger candidate.


In contrast, the model-based importance scores for each word $x_i$ is computed based on the hidden vectors of the models. In particular, we first form an overall feature vector $V_t$ that is used to predict the event type for $x_t$ via: 
$$V_t=[e_t,m^L_t, \text{max\_pool}(m^L_1, \cdots, m^L_n)]$$ 
In this work, we argue that the hidden vector of an important word in the sentence for ED should carry more useful information to predict the event type for $x_t$. Therefore, we consider a word $x_i$ as more important for the prediction of the trigger candidate $x_t$ if its representation $m^L_i$ is more similar to the vector $V_t$. We estimate the \textbf{model-based important scores} for every word $x_i$ with respect to the candidate $x_t$ as follow:
\begin{equation}
    q_i=\sigma(W^vV_t)\cdot \sigma(W^mm^L_i)
\end{equation}
where $W^v$ and $W^m$ are trainable parameters.

Afterward, we normalize the scores $P$ and $Q=q_1,\ldots,q_n$ using the softmax function. Finally, we minimize the KL divergence between the graph-based important scores $P$ and the model-based importance scores $Q$ by injecting a regularization term $\mathcal{L}_{ISC}$ (for the graph-model Importance Score Consistency) into the overall loss function:
\begin{equation}
\mathcal{L}_{ISC}(P,Q)=-\sum_{i=1}^n p_i\frac{p_i}{q_i}
\end{equation}
To predict the event type, we feed $V_t$ into a fully connected network with softmax function in the end to estimate the probability distribution $P(\hat{y}|X,t)$. To train the model, we use the negative log-likelihood as the classification loss $\mathcal{L}_{CE}=-\log P(y^*|X,t)$. Finally, we minimize the following combined loss function to train the proposed model:
\begin{equation}
\mathcal{L}=\mathcal{L}_{CE} + \alpha \mathcal{L}_{GD} +\beta \mathcal{L}_{ISC}
\end{equation}
where $\alpha$ and $\beta$ are trade-off coefficients.

\section{Experiments}
\label{sec:experiment}

\hspace{0.33cm} {\bf Datasets}: We evaluate our proposed model (called GatedGCN) on two ED datasets, i.e., ACE-2005 and Litbank. \textbf{ACE-2005} is a widely used benchmark dataset for ED, which consists of 33 event types. In contrast, \textbf{Litbank} is a newly published dataset in the literature domain, annotating words with two labels \textit{event} and \textit{none-event} \cite{sims2019literary}. Hence, on Litbank, we essentially solve trigger identification with a binary classification problem for the words. 

As the sizes of the ED dataset are generally small, the pre-processing procedures (e.g., tokenization, sentence splitting, dependency parsing, and selection of negative examples) might have a significant effect on the models' performance. 
For instance, the current best performance for ED on ACE-2005 is reported by \citep{yang2019exploring} (i.e., 80.7\% F1 score on the test set). However, once we re-implement this model and apply it to the data version pre-processed and provided by the prior work \citep{Nguyen:15:event,nguyen:18:graph}, we are only able to achieve an F1 score of 76.2\% on the test set. As the models share the way to split the data, we attribute such a huge performance gap to the difference in data pre-processing that highlights the need to use the same pre-processed data to measure the performance of the ED models. Consequently, in this work, we employ the exact data version that has been pre-processed and released by the early work on ED for ACE-2005 in \citep{Nguyen:15:event,nguyen:18:graph} and for Litbank in \cite{sims2019literary}.

The hyper-parameters for the models in this work are tuned on the development datasets, leading to the following selected values: one layer for the BiLSTM model with 128 hidden units in the layers, $L=2$ for the number of the GCN layers with 128 dimensions for the hidden vectors, 128 hidden units for the layers of all the feed-forward networks in this work, and $5e$-5 for the learning rate of the Adam optimizer. These values apply for both the ACE-2005 and Litbank datasets. For the trade-off coefficients $\alpha$ and $\beta$ in the overall loss function, we use $\alpha = 0.1$ and $\beta = 0.2$ for the ACE dataset while $\alpha=0.3$ and $\beta = 0.2$ are employed for Litbank. Finally, we use the case $\text{BERT}_\text{base}$ version of BERT and freeze its parameters during training in this work. To obtain the BERT representations of the word pieces, we use $M=12$ for ACE-2005 and $M=4$ for Litbank \cite{sims2019literary}.


\textbf{Results}: We compare our model with two classes of baselines on ACE-2005. The first class includes the models with  non-contextualized embedding, i.e., \textbf{CNN}: a CNN model \cite{Nguyen:15:event}, \textbf{NCNN}: non-consecutive CNN model: \cite{Nguyen:16b:modeling}, and \textbf{GCN-ED}: a GCN model \cite{nguyen:18:graph}. Note that these baselines use the same pre-processed data like ours. The second class of baselines concern the models with the contextualized embeddings, i.e., \textbf{DMBERT}: a model with dynamic pooling \cite{wang2019adversarial} and \textbf{BERT+MLP}: a MLP model with BERT \cite{yang2019exploring}. These models currently have the best-reported performance for ED on ACE-2005. Note that as these works employ different pre-processed versions of ACE-2005, we re-implement the models and tune them on our dataset version for a fair comparison. For Litbank, we use the following baselines reported in the original paper \cite{sims2019literary}: \textbf{BiLSTM}: a BiLSTM model with {\tt word2vec}, \textbf{BERT+BiLSTM}: a BiLSTM model with BERT, and \textbf{DMBERT} \cite{wang2019adversarial}.

\begin{table}[]
    \centering
     \resizebox{.4\textwidth}{!}{
    \begin{tabular}{l|c|c|c}
       \textbf{Model}  & \textbf{P} & \textbf{R} & \textbf{F} \\
       \hline
        CNN & 71.8 & 66.4 & 69.0 \\
        NCNN & - & - & 71.3 \\
        GCN-ED & 77.9 & 68.8 & 73.1 \\
       \hline
       DMBERT & 79.1 &	71.3 &	74.9 \\
       BERT+MLP & 77.8 &	74.6	& 76.2\\
       \hline
       GatedGCN (Ours) & 78.8 & 76.3 & \textbf{77.6} \\
    \end{tabular}
     }
    \caption{Performance on the ACE-2005 test set.}
    \label{tbl-ace}
\end{table}

\definecolor{a}{rgb}{1,0.3,0.3}
\definecolor{b}{rgb}{1,0.4,0.4}
\definecolor{c}{rgb}{1,0.6,0.6}
\definecolor{d}{rgb}{1,0.8,0.8}
\definecolor{e}{rgb}{1,0.96,0.96}
\newcommand{\hlc}[2]{{\sethlcolor{#1} \hl{#2}}}
\newcommand{\anchor}[1]{\underline{\textbf{#1}}}

\begin{figure*}[]
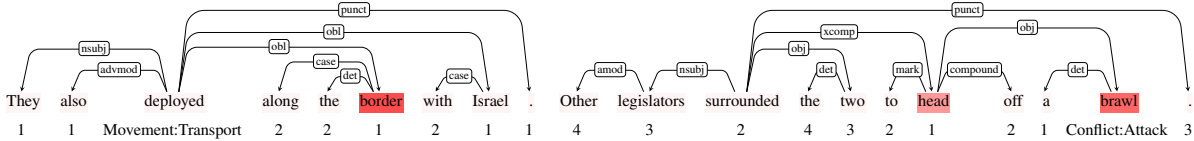

\centering
\resizebox{\textwidth}{!}{
\begin{dependency}[edge slant=6pt]
	\begin{deptext}[column sep=0.2cm, row sep=0.2cm]
		\hlc{e}{They} \& \hlc{e}{also} \& \hlc{e}{deployed} \& \hlc{e}{along} \& \hlc{e}{the} \& \hlc{a}{border} \& \hlc{e}{with} \& \hlc{e}{Israel} \& \hlc{e}{.} \\
		1\&1\& Movement:Transport \&2\&2\&1\&2\&1\&1 \\
	\end{deptext}
	\depedge{3}{1}{nsubj}
	\depedge{3}{2}{advmod}
	\depedge[edge height=6ex]{3}{6}{obl}
	\depedge[edge height=8ex]{3}{8}{obl}
	\depedge[edge height=11ex]{3}{9}{punct}
	\depedge[edge height=4ex]{6}{4}{case}
	\depedge[edge height=2ex]{6}{5}{det}
	\depedge[edge height=2ex]{8}{7}{case}
\end{dependency}
\begin{dependency}[edge slant=6pt]
	\begin{deptext}[column sep=0.2cm, row sep=0.2cm]
		\hlc{e}{Other} \& \hlc{e}{legislators} \& \hlc{e}{surrounded} \& \hlc{e}{the} \& \hlc{e}{two} \& \hlc{e}{to} \& \hlc{c}{head} \&[0.8cm] \hlc{e}{off} \& \hlc{e}{a} \& \hlc{b}{brawl} \& \hlc{e}{.} \\
		4\&3\&2\&4\&3\&2\&1\&2\&1\&Conflict:Attack\& 3\\
	\end{deptext}
	\depedge{2}{1}{amod}
	\depedge{3}{2}{nsubj}
	\depedge{3}{5}{obj}
	\depedge[edge height=8ex]{3}{7}{xcomp}
	\depedge[edge height=11ex]{3}{11}{punct}
	\depedge{5}{4}{det}
	\depedge{7}{6}{mark}
	\depedge{7}{10}{obj}
	\depedge[]{7}{8}{compound}
	\depedge[]{10}{9}{det}
\end{dependency}
}
\caption{Visualization of the model-based importance scores computed by the proposed model for several GatedGCN-successful examples. The words with bolder colors have larger importance scores in this case. Note that the golden event types ``\textit{Movement:Transport}" and ``\textit{Conflict:Attack}" are written under the trigger words in the sentences. Also, below each word in the sentences, we indicate the number of the words along the path from that word to the trigger word (i.e., the distances used in the graph-based importance scores). 
}
\label{fig-dep}
\end{figure*}

Table \ref{tbl-ace} presents the performance of the models on the ACE-2005 test set. This table shows that GatedGCN outperforms all the baselines with a significant improvement of 1.4\% F1-score over the second-best model BERT+MLP. In addition, Table \ref{tbl-litbank} shows the performance of the models on the Litbank test set. As can be seen, the proposed model is better than all the baseline models with 0.6\% F1-score improvement over the state-of-the-art model BERT+BiLSTM. These improvements are significant on both datasets ($p<0.05$), demonstrating the effectiveness of GatedGCN for ED.

\begin{table}[]
    \centering
     \resizebox{.42\textwidth}{!}{
    \begin{tabular}{l|c|c|c}
        \textbf{Model}  & \textbf{P} & \textbf{R} & \textbf{F} \\
        \hline
        BiLSTM & 70.4 & 60.7 & 65.2 \\
        + document context & 74.2 & 58.8 & 65.6 \\
        + sentence CNN & 71.6& 56.4 & 63.1 \\
        + subword CNN  & 69.2 & 64.8 & 66.9 \\
        \hline
        DMBERT & 65.0 &	76.7 & 70.4 \\
        BERT+BiLSTM & 75.5 & 72.3 & 73.9 \\
        \hline
        GatedGCN (Ours) & 69.9 & 79.8 & \textbf{74.5} \\
    \end{tabular}
     }
    \caption{Performance on the Litbank test set.}
    \label{tbl-litbank}
\end{table}

\textbf{Ablation Study}: The proposed model involves three major components: (1) the \textbf{Gates} to filter irrelevant information, (2) the Gate \textbf{Diversity} to encourage contextual distinction for the gates, and (3) the \textbf{Consistency} between graph and model-based importance scores. Table \ref{tbl-ablation-ace} reports the ablation study on the ACE-2005 development set when the components are incrementally removed from the full model (note that eliminating \textbf{Gate} also removes \textbf{Diversity} at the same time). As can be seen, excluding any component results in significant performance reduction, clearly testifying to the benefits of the three components in the proposed model for ED.

\textbf{Importance Score Visualization}: In order to further demonstrate the operation of the proposed model GatedGCN for ED, we analyze the model-based importance scores for the words in test set sentences of ACE-2005 that can be correctly predicted by GatedGCN, but leads to incorrect predictions for the ablated model ``-Gate-Consistency'' in Table \ref{tbl-ablation-ace} (called the GatedGCN-successful examples). In particular, Figure \ref{fig-dep} illustrates the model-based importance scores for the words in the sentences of several GatedGCN-successful examples. Among others, we find that although the trigger words are directly connected to several words (including the irrelevant ones) in these sentences, the {\bf Gates}, {\bf Diversity}, and {\bf Consistency} components in GatedGCN help to better highlight the most informative words among those neighboring words by assigning them larger importance scores. This enables the representation aggregation mechanism in GCN to learn better hidden vectors, leading to improved performance for ED in this case.


\begin{table}[h!]
    \centering
     \resizebox{.48\textwidth}{!}{
    \begin{tabular}{l|c|c|c}
       \textbf{Model}  & \textbf{P} & \textbf{R} & \textbf{F} \\
       \hline
       GatedGCN (full) & 76.7 & 70.5& \textbf{73.4} \\
       \hline
       -Diversity &78.5	& 67.0& 72.3 \\
        -Consistency & 80.5 & 64.7&  71.7 \\
        -Diversity -Consistency & 79.0 & 63.0 &  70.1 \\
        \hline
        -Gates & 77.8 &	65.3 & 71.3\\
        -Gates -Consistency & 83.0 &	62.5 & 71.0\\
    \end{tabular}
     }
    \caption{Ablation study on the ACE-2005 dev set.}
    \label{tbl-ablation-ace}
\end{table}

\section{Conclusion}

We demonstrate how gating mechanisms, gate diversity, and graph structure can be used to integrating syntactic information and improve the hidden vectors for ED models. The proposed model achieves state-of-the-art performance on two ED datasets. In the future, we plan to apply the proposed model for the related tasks and other settings of ED, including new type extension \cite{Nguyen:16b,Lai:19}, and few-shot learning \cite{Lai:20a,Lai:20b}.


\section*{Acknowledgement}


This research has been supported in part by Vingroup Innovation Foundation (VINIF) in project code VINIF.2019.DA18 and Adobe Research Gift. This research is also based upon work supported in part by the Office of the Director of National Intelligence (ODNI), Intelligence Advanced Research Projects Activity (IARPA), via IARPA Contract No. 2019-19051600006 under the Better Extraction from Text Towards Enhanced Retrieval (BETTER) Program. The views and conclusions contained herein are those of the authors and should not be interpreted as necessarily representing the official policies, either expressed or implied, of ODNI, IARPA, the Department of Defense, or the U.S. Government. The U.S. Government is authorized to reproduce and distribute reprints for governmental purposes notwithstanding any copyright annotation therein. This document does not contain technology or technical data controlled under either the U.S. International Traffic in Arms Regulations or the U.S. Export Administration Regulations.

\bibliography{ref}

\begin{thebibliography}{31}
\expandafter\ifx\csname natexlab\endcsname\relax\def\natexlab#1{#1}\fi

\bibitem[{Ahn(2006)}]{Ahn:06}
David Ahn. 2006.
\newblock The stages of event extraction.
\newblock In \emph{Proceedings of the Workshop on Annotating and Reasoning
  about Time and Events}.

\bibitem[{Chen et~al.(2017)Chen, Liu, Zhang, Liu, and Zhao}]{Chen:17}
Yubo Chen, Shulin Liu, Xiang Zhang, Kang Liu, and Jun Zhao. 2017.
\newblock Automatically labeled data generation for large scale event
  extraction.
\newblock In \emph{ACL}.

\bibitem[{Chen et~al.(2015)Chen, Xu, Liu, Zeng, and Zhao}]{Chen:15}
Yubo Chen, Liheng Xu, Kang Liu, Daojian Zeng, and Jun Zhao. 2015.
\newblock Event extraction via dynamic multi-pooling convolutional neural
  networks.
\newblock In \emph{ACL-IJCNLP}.

\bibitem[{Chen et~al.(2018)Chen, Yang, Liu, Zhao, and Jia}]{Chen:18}
Yubo Chen, Hang Yang, Kang Liu, Jun Zhao, and Yantao Jia. 2018.
\newblock Collective event detection via a hierarchical and bias tagging
  networks with gated multi-level attention mechanisms.
\newblock In \emph{EMNLP}.

\bibitem[{Devlin et~al.(2019)Devlin, Chang, Lee, and
  Toutanova}]{devlin2019bert}
Jacob Devlin, Ming-Wei Chang, Kenton Lee, and Kristina Toutanova. 2019.
\newblock Bert: Pre-training of deep bidirectional transformers for language
  understanding.
\newblock In \emph{NAACL:HLT}, pages 4171--4186.

\bibitem[{Feng et~al.(2016)Feng, Huang, Tang, Ji, Qin, and Liu}]{Feng:16}
Xiaocheng Feng, Lifu Huang, Duyu Tang, Heng Ji, Bing Qin, and Ting Liu. 2016.
\newblock A language-independent neural network for event detection.
\newblock In \emph{ACL (Volume 2: Short Papers)}, volume~2, pages 66--71.

\bibitem[{Hong et~al.(2011)Hong, Zhang, Ma, Yao, Zhou, and Zhu}]{Hong:11}
Yu~Hong, Jianfeng Zhang, Bin Ma, Jianmin Yao, Guodong Zhou, and Qiaoming Zhu.
  2011.
\newblock Using cross-entity inference to improve event extraction.
\newblock In \emph{ACL}.

\bibitem[{Jagannatha and Yu(2016)}]{Jagannatha:16}
Abhyuday~N Jagannatha and Hong Yu. 2016.
\newblock Bidirectional rnn for medical event detection in electronic health
  records.
\newblock In \emph{NAACL}.

\bibitem[{Ji and Grishman(2008)}]{Ji:08}
Heng Ji and Ralph Grishman. 2008.
\newblock Refining event extraction through cross-document inference.
\newblock In \emph{ACL}.

\bibitem[{Kipf and Welling(2017)}]{Kipf:17}
Thomas~N. Kipf and Max Welling. 2017.
\newblock Semi-supervised classification with graph convolutional networks.
\newblock In \emph{ICLR}.

\bibitem[{Lai et~al.(2020{\natexlab{a}})Lai, Dernoncourt, and Nguyen}]{Lai:20a}
Viet~Dac Lai, Franck Dernoncourt, and Thien~Huu Nguyen. 2020{\natexlab{a}}.
\newblock Exploiting the matching information in the support set for few shot
  event classification.
\newblock In \emph{Proceedings of the 24th Pacific-Asia Conference on Knowledge
  Discovery and Data Mining (PAKDD)}.

\bibitem[{Lai et~al.(2020{\natexlab{b}})Lai, Dernoncourt, and Nguyen}]{Lai:20b}
Viet~Dac Lai, Franck Dernoncourt, and Thien~Huu Nguyen. 2020{\natexlab{b}}.
\newblock Extensively matching for few-shot learning event detection.
\newblock In \emph{Proceedings of the 1st Joint Workshop on Narrative
  Understanding, Storylines, and Events (NUSE) at ACL 2020}.

\bibitem[{Lai and Nguyen(2019)}]{Lai:19}
Viet~Dac Lai and Thien Nguyen. 2019.
\newblock Extending event detection to new types with learning from keywords.
\newblock In \emph{Proceedings of the 5th Workshop on Noisy User-generated Text
  (W-NUT 2019) at EMNLP 2019}.

\bibitem[{Li et~al.(2013)Li, Ji, and Huang}]{Li:13}
Qi~Li, Heng Ji, and Liang Huang. 2013.
\newblock Joint event extraction via structured prediction with global
  features.
\newblock In \emph{ACL}.

\bibitem[{Liu et~al.(2017)Liu, Chen, Liu, and Zhao}]{Liu:17}
Shulin Liu, Yubo Chen, Kang Liu, and Jun Zhao. 2017.
\newblock Exploiting argument information to improve event detection via
  supervised attention mechanisms.
\newblock In \emph{ACL}.

\bibitem[{Liu et~al.(2018)Liu, Luo, and Huang}]{liu2018jointly}
Xiao Liu, Zhunchen Luo, and He-Yan Huang. 2018.
\newblock Jointly multiple events extraction via attention-based graph
  information aggregation.
\newblock In \emph{EMNLP}, pages 1247--1256.

\bibitem[{Marcheggiani and Titov(2017)}]{Marcheggiani:17}
Diego Marcheggiani and Ivan Titov. 2017.
\newblock Encoding sentences with graph convolutional networks for semantic
  role labeling.
\newblock In \emph{EMNLP}.

\bibitem[{Mitamura et~al.(2015)Mitamura, Liu, and Hovy}]{Mitamura:15}
Teruko Mitamura, Zhengzhong Liu, and Eduard Hovy. 2015.
\newblock Overview of tac kbp 2015 event nugget track.
\newblock In \emph{TAC}.

\bibitem[{Nguyen et~al.(2016)Nguyen, Cho, and Grishman}]{Nguyen:16a:joint}
Thien~Huu Nguyen, Kyunghyun Cho, and Ralph Grishman. 2016.
\newblock Joint event extraction via recurrent neural networks.
\newblock In \emph{NAACL}.

\bibitem[{Nguyen et~al.(2016b)Nguyen, Fu, Cho, and Grishman}]{Nguyen:16b}
Thien~Huu Nguyen, Lisheng Fu, Kyunghyun Cho, and Ralph Grishman. 2016b.
\newblock A two-stage approach for extending event detection to new types via
  neural networks.
\newblock In \emph{Proceedings of the 1st ACL Workshop on Representation
  Learning for NLP (RepL4NLP)}.

\bibitem[{Nguyen and Grishman(2015)}]{Nguyen:15:event}
Thien~Huu Nguyen and Ralph Grishman. 2015.
\newblock Event detection and domain adaptation with convolutional neural
  networks.
\newblock In \emph{ACL-IJCNLP}.

\bibitem[{Nguyen and Grishman(2016)}]{Nguyen:16b:modeling}
Thien~Huu Nguyen and Ralph Grishman. 2016.
\newblock Modeling skip-grams for event detection with convolutional neural
  networks.
\newblock In \emph{EMNLP}.

\bibitem[{Nguyen and Grishman(2018)}]{nguyen:18:graph}
Thien~Huu Nguyen and Ralph Grishman. 2018.
\newblock Graph convolutional networks with argument-aware pooling for event
  detection.
\newblock In \emph{AAAI}.

\bibitem[{Nguyen et~al.(2016g)Nguyen, Meyers, and Grishman}]{Nguyen:16g}
Thien~Huu Nguyen, Adam Meyers, and Ralph Grishman. 2016g.
\newblock New york university 2016 system for kbp event nugget: A deep learning
  approach.
\newblock In \emph{Proceedings of Text Analysis Conference (TAC)}.

\bibitem[{Nguyen and Nguyen(2019)}]{Nguyen:19}
Trung~Minh Nguyen and Thien~Huu Nguyen. 2019.
\newblock One for all: Neural joint modeling of entities and events.
\newblock In \emph{AAAI}.

\bibitem[{Sims et~al.(2019)Sims, Park, and Bamman}]{sims2019literary}
Matthew Sims, Jong~Ho Park, and David Bamman. 2019.
\newblock Literary event detection.
\newblock In \emph{Proceedings of the 57th Annual Meeting of the Association
  for Computational Linguistics}, pages 3623--3634.

\bibitem[{Veyseh et~al.(2019)Veyseh, Nguyen, and Dou}]{veyseh:19b}
Amir Pouran~Ben Veyseh, Thien~Huu Nguyen, and Dejing Dou. 2019.
\newblock Graph based neural networks for event factuality prediction using
  syntactic and semantic structures.
\newblock In \emph{ACL}.

\bibitem[{Walker et~al.(2006)Walker, Strassel, Medero, and
  Maeda}]{walker2006ace}
Christopher Walker, Stephanie Strassel, Julie Medero, and Kazuaki Maeda. 2006.
\newblock Ace 2005 multilingual training corpus.
\newblock \emph{Linguistic Data Consortium, Philadelphia}, 57.

\bibitem[{Wang et~al.(2019)Wang, Han, Liu, Sun, and Li}]{wang2019adversarial}
Xiaozhi Wang, Xu~Han, Zhiyuan Liu, Maosong Sun, and Peng Li. 2019.
\newblock Adversarial training for weakly supervised event detection.
\newblock In \emph{NAACL-HLT}, pages 998--1008.

\bibitem[{Yan et~al.(2019)Yan, Jin, Meng, Guo, and Cheng}]{yan2019event}
Haoran Yan, Xiaolong Jin, Xiangbin Meng, Jiafeng Guo, and Xueqi Cheng. 2019.
\newblock Event detection with multi-order graph convolution and aggregated
  attention.
\newblock In \emph{EMNLP-IJCNLP}, pages 5770--5774.

\bibitem[{Yang et~al.(2019)Yang, Feng, Qiao, Kan, and Li}]{yang2019exploring}
Sen Yang, Dawei Feng, Linbo Qiao, Zhigang Kan, and Dongsheng Li. 2019.
\newblock Exploring pre-trained language models for event extraction and
  generation.
\newblock In \emph{Proceedings of the 57th Annual Meeting of the Association
  for Computational Linguistics}, pages 5284--5294.

\end{thebibliography}
\bibliographystyle{acl_natbib}

\end{document}